# Towards dialect-inclusive recognition in a low-resource language: are balanced corpora the answer?

*[1]Liam Lonergan, [2]Mengjie Qian, [1]Neasa Ní Chiaráin, [1]Christer Gobl, [1]Ailbhe Ní Chasaide*

[1]Phonetics and Speech Laboratory, School of Linguistics, Speech and Communication Sciences, Trinity College Dublin, Ireland
[2]Engineering Department, Cambridge University, UK
[2]mq227@cam.ac.uk, [1]{llonerga, nichiarn, cegobl, anichsid}@tcd.ie

## Abstract

ASR systems are generally built for the spoken 'standard', and their performance declines for non-standard dialects/varieties. This is a problem for a language like Irish, where there is no single spoken standard, but rather three major dialects: Ulster (Ul), Connacht (Co) and Munster (Mu). As a diagnostic to quantify the effect of the speaker's dialect on recognition performance, 12 ASR systems were trained, firstly using baseline dialect-balanced training corpora, and then using modified versions of the baseline corpora, where dialect-specific materials were either subtracted or added. Results indicate that dialect-balanced corpora do not yield a similar performance across the dialects: the Ul dialect consistently underperforms, whereas Mu yields lowest WERs. There is a close relationship between Co and Mu dialects, but one that is not symmetrical. These results will guide future corpus collection and system building strategies to optimise for cross-dialect performance equity.

**Index Terms**: bias, speech recognition, low-resource, multi-dialect ASR, Irish Gaelic.

## 1. Introduction

Speech technology is a ubiquitous feature in today's world. Of the world's 6,000+ languages, only less than 1% have sufficient speech corpora for speech technology development [1]. This has resulted in an unbalanced linguistic landscape, where the few languages for which such technologies have been sufficiently developed, determine the linguistic means through which people interact with voice-assisted AI technologies.

For the endangered or under-resourced language, major dialect variation compounds the problem of data sparsity. In the 'major' languages, speech technologies are initially developed for the standard dialect and non-standard varieties are typically added later. This is not an option for Irish, which like many minority and endangered languages, does not have a spoken standard, but rather, three different dialects: Ulster (Ul), Connacht (Co) and Munster (Mu) as illustrated in Figure 1.

The issues of bias and inclusivity thus need to be considered from the outset, and have been central in the ABAIR initiative, which has been developing speech technology for Irish (https://abair.ie/). TTS systems have been developed for each of the three dialect groups and are available on the website. The adequacy of ABAIR's applications that incorporate speech technology (e.g. language learning platforms [2], assistive technologies [3]) depends on their availability in dialect appropriate forms. In this context, equity of cross-dialect performance is a core aspiration of ABAIR's ASR development. Developing multi-dialect technology presents many challenges for a low-resource language, but it is also an opportunity for technology to capture the richness and diversity of the living language. An understanding of the linguistic structure and a sensitivity to the sociolinguistic context in which the technology will be deployed is paramount. This outlook along with a recognition of the importance of speech and language technologies to the preservation of the language are formally articulated in the *Digital Plan for the Irish Language* [4] as part of the *20 Year Strategy for the Irish Language 2010 – 2030* [5].

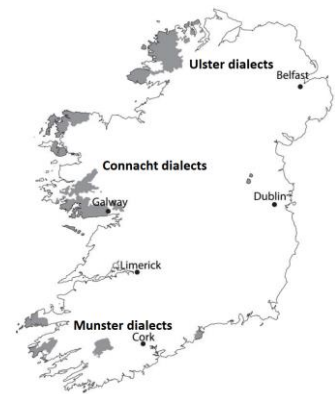

Figure 1: *Map showing the Gaeltacht dialect regions of Ulster (UL), Connacht (CO) and Munster (MU).*

To date, corpus development efforts for the current ABAIR ASR system have strived for balance among the dialects as far as possible. Nonetheless, past experiments have indicated that a dialect-balanced corpus does not bring equal performance across dialects. Therefore, the aim of this paper is to explore the performance outcomes for individual dialects, by training a variety of systems, where an initial, dialect-balanced training corpus is modified systematically by either subtracting or supplementing dialect-specific corpora, one dialect at a time. The ASR system employed in these experiments is an XLS-R wav2vec 2.0 model [6], finetuned with Connectionist Temporal Classification (CTC), and is described in more detail below. This study therefore aims to:

(i) **Extend our understanding of cross-dialect distance**.

(ii) **Serve as a diagnostic test to guide corpus collection** so as to maximise cross-dialect performance equity. Corpora are currently being collected using the crowdsourcing facility *MíleGlór* (see website) and with field recordings.

(iii) **Inform dialect-bias mitigation strategies** for future iterations of the ABAIR ASR system for Irish.

## 2. Background

### 2.1 The Irish language

Irish is an endangered [7] Celtic language and is spoken as a community language in relatively small Gaeltacht regions, mostly located in remote parts of the western seaboard (see Figure 1). Over centuries, colonialization, the defeat of local aristocracy, and the consequent collapse of the Gaelic order all led to increasing language attrition. Decline during the $19^{th}$ century was accelerated by the great famine and its aftermath of mass emigration. With the westwards drift of language loss, the Irish-speaking communities increasingly became linguistic 'islands' with relatively little communication between them. More recently, the advent of Irish language radio and television has led to greater cross-dialect exposure and intelligibility [8].

The dialects differ in aspects of their pronunciation, prosody, lexicon, morphology, and syntax. Although not quantified, linguists have tended to have differing views regarding the groupings. O'Rahilly [9] posits only two dialects, a northern grouping (UL) and a southern grouping (CO and MU). Traditional varieties are classically described [9] as being on a geographic dialect continuum with poles in the northeast and south where different innovations started in the late medieval to early modern period and gradually exerted influences south-westwards and northwards respectively [10]. Some recent comparative studies of dialect intonation [11, 12, 13] do suggest a north–south divide, but in terms of stress patterns, the MU dialect stands out [14]. It should be noted however, that O Rahilly's analysis of two (northern and southern) groupings is not generally accepted. The general consensus is that Irish has three main dialect groups [15, 16] – as presented here in the Introduction.

### 2.2 Low-resource ASR and dialect bias

Where dialects are different enough from each other and enough data is available for each, it is common practice to build ASR models for each dialect separately [17, 18, 19]. However, if there is only limited data available, pooling each dialect's resources together has also proven beneficial [20, 21]. Despite the considerable differences between the dialects of Irish, the low-resource context i.e., a lack of large parallel speech-text corpora, has determined a strategy of pooling all dialect materials whilst striving to maintain a cross-dialect balance. As mentioned however, current results indicate unequal dialect performance outcomes.

Quantifying dialect and accent bias is an emerging topic of inquiry. A clear bias is demonstrated in [22], where a commercial ASR system for English was shown to perform worse for Scottish speakers of English than for speakers from the United States or New Zealand. An inherent racial bias is reported in [23] for five popular, commercial ASR services for US speakers of English, where it is found that word error rates for black speakers are nearly twice as large as for white speakers in every case.

The impact of inadequate dialect coverage is also attracting attention. A follow-up study to [23] explores the psychological and experiential effect of such racially-biased ASR systems [24], and found that recognition errors negatively impacted African American users, leaving them feeling othered and with the feeling that the technology was not made for them. As a result, these users accommodate their speech to have better success with the technology. Therefore, when designing a recognition system for a new language, it is important to consider how society and individuals will interact with the technology, to ensure that societal values, such as equitable recognition performance for speakers of different dialects, are embedded in the system. Such a sociotechnical perspective [25] is, as mentioned earlier, a primary concern for ABAIR's Irish ASR systems.

Strategies to mitigate dialect and accent bias in ASR include the balancing of training corpora [26], and the explicit modelling of dialect or accent in the ASR process. The present experiments are focused on the former strategy, but the latter approach is revisited in the Discussion.

### 2.3 ABAIR's Irish ASR system: progress to date

ABAIR's work carried out on Irish ASR has to date focused on modular systems, which due to the limited amount of available paired audio-text data for Irish, consistently outperform systems trained in an End-to-End (E2E) manner. Initial work looked at improving the lexicon and language model components [27], which gave the first glimpses into Irish ASR systems being susceptible to dialect bias. This was followed by a closer inspection of the lexicon [28] and an attempt to find an optimal solution to lexicon building for cross-dialect variation. Through this experiment, a Global approach adopting dialect-invariant phonemes/morphemes where the dialects are known to diverge in pronunciation, proved to be most successful. Special attention was given to ensure a balance of dialects in the training and test sets. Nonetheless, the results clearly showed a preference for the Munster dialect over the other two for all lexicon set-ups. The best performing system in [28] achieved a word-error-rate (WER) of 8.74% overall, and is publicly accessible via the website (https://abair.ie/speech-recognition).

## 3. Methods

Indications of dialect bias from the earlier studies prompted the current work, which sets out to explore how the composition of a training corpus affects performance for individual dialects. To this end, dialect-balanced training sets are systematically modified by subtracting or supplementing dialect-specific materials, resulting in 12 training sets. Each training set is used to finetune an XLS-R pretrained model with CTC. These systems are evaluated using a dialect-balanced evaluation set and cross-dialect performance outcomes are analysed.

### 3.1 Data: Training set modifications

Details for the training sets used are given in Table 1 alongside the results that are discussed in Section 4.

**Experiment 1:** For this experiment, 16 hours of read speech from each dialect are used, giving a combined total of 48 hours. E2E systems such as the one employed here perform poorly when the training set is too small, and as Experiment 1 involves the removal of full dialect datasets at a time (which would result in a training set size of 32 hours), an extra corpus of non-native speech (5 hours) was included to bulk up the training data in all training sets, with a view to increasing the reliability of the results. This could introduce inadvertent bias, as non-native speech can to varying degrees approximate one of the dialects, and this question is addressed below. This results in 4 training sets, the baseline *B small* (53h), and reduced sets *-UL* (37.9h), *-CO* (36.7h), and *-MU* (37.2h). Additionally, to ensure a fair comparison between the datasets with a dialect removed (and a dialect-balanced baseline), the initial *B small* set was resized, maintaining the 5h of non-native speech while reducing the proportion of each dialect from 16h to 11h. The resized baseline *B small res* was 37.5h.

**Experiment 2:** To investigate the effect on performance when the representation of one dialect is increased, 30h of data from each dialect were collected from a corpus of spontaneous speech. This is quite different from the read speech, and contained mostly conversational, broadcast materials. Previous experiments indicated that adding such data is particularly helpful for data-intensive E2E approaches, even if the data is not of the target domain (i.e. read speech). This yielded 3 more training sets: *+UL* (83.5h), *+CO* (83.5h), and *+MU* (83.5h). In addition to these training sets, a further baseline is built, combining all three dialect-specific spontaneous corpora with all read data: *B large* (143.5h).

**Experiment 3:** This involved a more extreme version of Experiment 1, where one dialect dataset is removed at a time. The starting point was the combined dataset *B large* (143.5h), which contained both the read and spontaneous data for each of the 3 dialects. From this, entire individual dialect datasets were removed, yielding the following: *--UL* (97.9h), *--CO* (96.7h), *--MU* (97.2h).

### 3.2 Data: test sets

The development (1.5h) and evaluation (2.5h) test sets involved speakers and utterance texts not included in the training set, and these were balanced for dialect and for gender (within dialect). The development set was used during training and the evaluation set is used to evaluate the systems, with results shown in Table 1 and Figure 2.

### 3.3 System description

The ASR system employed in the present experiments is the 300 million parameter wav2vec 2.0 – XLS-R model, which is finetuned with CTC using the fairseq framework [29]. Due to the assumptions of conditional independence that underpin CTC, the resulting ASR systems are effectively just acoustic models with a shallow character prediction history, which often leads to sub-optimal performance [30]. To overcome this, a language model trained on text-only data is often incorporated into decoding. However, available Irish-language text corpora are not tagged with dialect information, therefore it is unclear to what extent they contain dialect biases of their own. As such, incorporating an external language model could lead to an unintended preference for one dialect over the others, that could be difficult to disentangle in the analysis. As the aim of this paper is to analyse potential dialect biases that could be introduced by the proportional representation of the dialects in a training set, no external language model was used here. Consequently, the WERs reported are quite high, but the interest here is rather on the difference in performance outcomes across the dialects, and the relative performance improvements and degradations that emerge with the training set modifications.

## 4. Results

The results for Experiments 1, 2 and 3 are displayed in Table 1, where the duration for each training set is provided, alongside the overall WER for the evaluation set, in addition to a breakdown of this result by speakers' dialect. This table is sectioned into blocks for each experiment, where the relevant baseline for comparison (blue rows) is shown at the top of each block. A further drilldown of these results is displayed in panels A and B of Figure 2: panel A shows the WER results by dialect for each experiment alongside the relevant baseline for comparison; panel B shows for each dialect the relative WERs to the baselines of each experiment.

**Results for baseline systems:** When a system is trained with one of the dialect-balanced, baseline training sets, there is a clear, consistent difference in performance outcomes for the dialects. In all cases, the performance for Ul speakers is considerably worse than the other dialects (see *B small*, *B small res* and *B large* in Panel A, Figure 2). Performance for Mu speakers emerges as the best. The difference in performance between Co and Mu speakers is clearly marked in both *B small* and in *B large*, but negligible for the limited dataset *B small res*.

As the cross-dialect performance trends are consistent across the *B small* and the more extensive *B large* corpora, and given that the 5hr of non-native speech represents a minute portion of *B large*, it can be inferred that the inclusion of non-native speech is not likely to be skewing results.

Table 1: *Training sets and results for experiments 1, 2 and 3.*

| Training sets | | Results - WER | | | |
|---|---|---|---|---|---|
| Training set | duration | all speakers | Ul speakers | Co speakers | Mu speakers |
| B small | 53.5h | 33.8 | 37.4 | 32.5 | 30.7 |
| **Experiment 1: removes 1 dialect from B small** | | | | | |
| B small res | 37.45h | 36.4 | 39.7 | 34.2 | 34.1 |
| - UL | 37.9h | 41.7 | 52.7 | 36.0 | 33.0 |
| - CO | 36.7h | 38.0 | 40.5 | 39.1 | 34.7 |
| - MU | 37.2h | 42.2 | 40.5 | 38.1 | 46.2 |
| **Experiment 2: adds 30h per dialect to B small** | | | | | |
| B small | 53.5h | 33.8 | 37.4 | 32.5 | 30.7 |
| + UL | 83.5h | 28.5 | 31.0 | 28.6 | 25.8 |
| + CO | 83.5h | 30.4 | 34.9 | 28.8 | 26.3 |
| + MU | 83.5h | 29.1 | 34.5 | 28.2 | 23.6 |
| **Experiment 3: removes 1 dialect from B large** | | | | | |
| B large | 143.5h | 23.4 | 26.3 | 23.7 | 20.0 |
| -- UL | 97.9h | 33.5 | 46.1 | 28.0 | 22.8 |
| -- CO | 96.7h | 27.7 | 29.4 | 30.5 | 24.3 |
| -- MU | 97.2h | 31.7 | 29.2 | 29.2 | 35.8 |

### 4.1 Experiment 1 & 3: subtracting dialect-specific datasets

When a dialect-specific dataset is removed, the negative effects on performance for Ul and Mu speakers are quite extreme (Figure 2). For Co speakers, there is relatively little degradation in performance by the removal of its dataset – in stark contrast to the effects in the other dialects. In Panel B, the relatively flat line for Co (showing little effect, regardless of which dataset is removed), contrasts strikingly with the major shifts seen for the other two dialects in both Experiments 1 and 3.

There is a striking asymmetry between performance for Co and Mu speakers with the removal of their respective datasets. For Co speakers, performance is roughly similar when CO or MU dialect-specific datasets are removed. However, the performance for Mu speakers is much more severely affected by the removal of the MU dataset, than by the removal of the CO dataset.

### 4.2 Experiment 2: supplementing dialect-specific datasets

Supplementing data improves the performance for each dialect, but again, cross-dialect differences emerge. Panel A of Figure 2 suggests that the performance for Co speakers (green) is improved roughly equally regardless of the dataset that is added. Performance for Ul and Mu speakers is however more enhanced when the supplemented data is dialect appropriate.

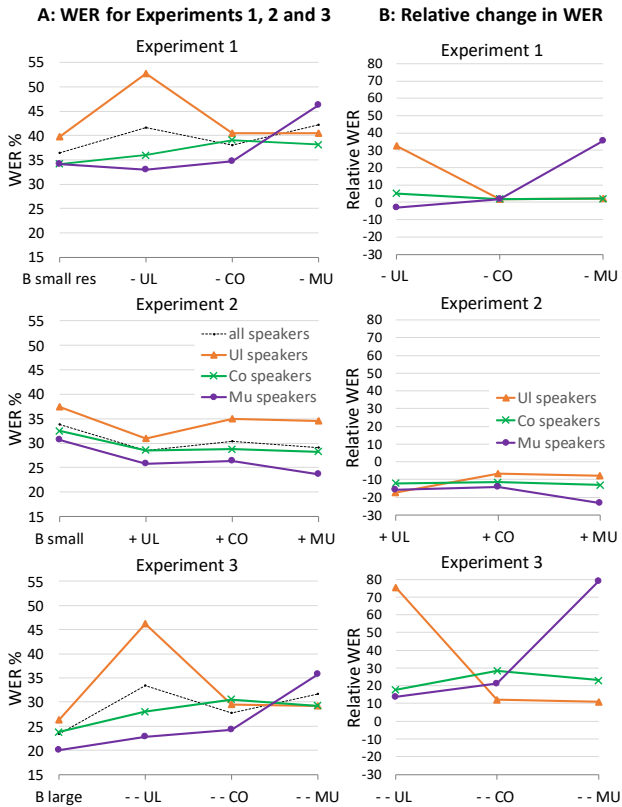

Figure 2: *Panel A - WER results for all three experiments. Panel B - change in WER relative to the baseline.*

## 5. Discussion

**(i) Extending our understanding of cross-dialect distance:** This study casts light on the linguistic diversity of the Irish dialects and their relative acoustic proximity. The results suggest a gap between the Ul dialect and the other two. This would support O'Rahilly's proposition of a north (Ul) – south (Co and Mu) divide. Note however, that the models employed in these experiments capture largely segmental aspects of speech and exclude important dialect markers such as differences in prosody, morphology, lexicon, and syntax. Furthermore, there are striking differences between the Southern Co and Mu dialects. For Co, performance is roughly equally enhanced by the addition of any dataset, whereas for Mu (and Ul), dialect-specific data is critical. In this sense, Co emerges as a "mid" dialect – matching its geographical location and this could be construed as supporting O'Rahilly's suggestion [9] of a continuum of dialects with poles in the north and south with Co being somewhere between them. Nonetheless, the consensus of three distinct dialects does appear to be warranted.

**(ii) Serving as a diagnostic test to guide corpus collection:** These experiments set out to clarify cross-dialect recognition biases to guide the ongoing process of ABAIR corpus development for Irish ASR. Balanced corpora are viewed as an ideal solution for mitigating bias [26]. The results here suggest that this is only partially successful: the balanced corpora yield considerably poorer WERs for Ul speakers, and best WERs for Mu speakers with Co falling between the two. The corpora used here are small in scale, however the consistent trend in performance disparity with balanced sets indicates this trend will be borne out with corpora. These results suggest that differential dialect weightings at the point of corpus collection could be a possible bias-mitigating strategy. Going forward, the focus will be on Ul materials, to bridge the performance gap. Beyond that, some positive weighting towards Mu might also be warranted, given that it should benefit both Mu and Co recognition, while the reverse is not the case. Curiously, while one might have expected that adding data from a "mid" dialect might confer the most benefit overall, the results here suggest the opposite.

In practical terms, and particularly in a low-resource context, the ideal of balanced or weighted corpora may not be an option. Furthermore, depending on the availability of corpora, pooling all data may produce the best possible system. The diagnostic tests used here are not proposed with the intention that one should limit training sets, but rather as a useful strategy to assess and monitor the dialect biases in emerging systems. Depending on the language and developmental context, such tests might guide adjustments to corpus collection strategies as is intended here for Irish. But even where such corpus adjustments are not an option, the results of such tests can also inform further strategies to alleviate dialect bias.

**(iii) Informing dialect-bias mitigation strategies:** There are many dialect-bias mitigation strategies that can be adopted, which involve explicitly modelling dialect in the ASR pipeline. One approach is to either prepend or append dialect labels to the target utterance text, thereby forcing the system to learn dialect-specific characteristics [30, 31]. Alternatively, 1-hot dialect vectors or dialect embeddings can be used as auxiliary input features to ASR systems, something which has been demonstrated to improve recognition performance [33]. Yet a further approach is to jointly train part of the ASR system to perform dialect identification, such that the mutual information between this task and ASR can be leveraged to improve recognition performance for dialects [33, 34]. Additionally, the inclusion of categorical, demographic information, including dialect, has been shown to improve recognition performance in a multi-task learning set-up [36].

Current research is exploring how these strategies of incorporating dialect information can reduce the disparity in dialect performance for Irish and improve recognition overall.

## 6. Conclusions

This study reveals that even with dialect-balanced corpora, there can be considerable dialect biases in ASR performance. The diagnostic test used here are proposed as a useful device that could be periodically carried out to monitor the bias and to quantify the success in overcoming it. The study provides further insight into dialect divergence and can act as a relatively impartial method that could complement more phonetic and linguistic classification. Mitigating dialect bias by corpus adjustments or by explicitly modelling dialect as part of the ASR process, are one aspect of a wider issue. It is critical that technology development is sensitive to the language, its dialects, and the sociolinguistic context of its speaker communities.

## 7. Acknowledgement

The ABAIR initiative is supported by the Department of Tourism, Culture, Art, Gaeltacht, Sport & Media, with funding from the National Lottery, as part of the 20 Year Plan for Irish 2010-30.